\pdfoutput=1

\documentclass[11pt]{article}

\usepackage[]{ACL2023}

\usepackage{times}
\usepackage{latexsym}

\usepackage[T1]{fontenc}

\usepackage[utf8]{inputenc}

\usepackage{microtype}

\usepackage{inconsolata}

\usepackage{graphicx}
\usepackage{tabularx}
\usepackage{makecell}
\usepackage{tikz}
\usepackage{enumitem}
\usepackage[]{algorithm2e}
\usepackage{listings}
\usepackage{float}
\usepackage{amsmath}

\usepackage{blindtext}
\newcommand\blfootnote[1]{%
\begingroup
\renewcommand\thefootnote{}\footnote{#1}%
\addtocounter{footnote}{-1}%
\endgroup
}

\title{README: Bridging Medical Jargon and Lay Understanding for 

Patient Education through Data-Centric NLP}

\author{Zonghai Yao $^1$, Nandyala Siddharth Kantu $^1$, Guanghao Wei $^1$, Hieu Tran $^1$, \\
\bf{Zhangqi Duan} $^1$, \bf{Sunjae Kwon} $^1$, \bf{Zhichao Yang}$^1$, \bf{README annotation team}$^{2}$, \bf{Hong Yu}$^{1, 2}$\\
University of Massachusetts, Amherst$^1$, University of Massachusetts, Lowell$^2$\\
{\tt zonghaiyao@umass.edu}\\
}

\begin{document}
\maketitle
\begin{abstract}
The advancement in healthcare has shifted focus toward patient-centric approaches, particularly in self-care and patient education, facilitated by access to Electronic Health Records (EHR). However, medical jargon in EHRs poses significant challenges in patient comprehension. To address this, we introduce a new task of automatically generating lay definitions, aiming to simplify complex medical terms into patient-friendly lay language. 
We first created the README dataset, an extensive collection of over 50,000 unique (medical term, lay definition) pairs and 300,000 mentions, each offering context-aware lay definitions manually annotated by domain experts. We have also engineered a data-centric Human-AI pipeline that synergizes data filtering, augmentation, and selection to improve data quality. We then used README as the training data for models and leveraged a Retrieval-Augmented Generation method to reduce hallucinations and improve the quality of model outputs. Our extensive automatic and human evaluations demonstrate that open-source mobile-friendly models, when fine-tuned with high-quality data, are capable of matching or even surpassing the performance of state-of-the-art closed-source large language models like ChatGPT. This research represents a significant stride in closing the knowledge gap in patient education and advancing patient-centric healthcare solutions ~\footnote{Our data is released at \url{https://github.com/seasonyao/NoteAid-README} and \url{https://huggingface.co/datasets/bio-nlp-umass/NoteAid-README} with CC-BY-NC 4.0 license.}.
\end{abstract}

\section{Introduction}
\label{Intro}

\blfootnote{$\dagger$ To appear in Findings of the Association for Computational Linguistics: EMNLP 2024}

\begin{figure}
    \centering
    \vspace{-2mm}
    \includegraphics[width=1\linewidth]{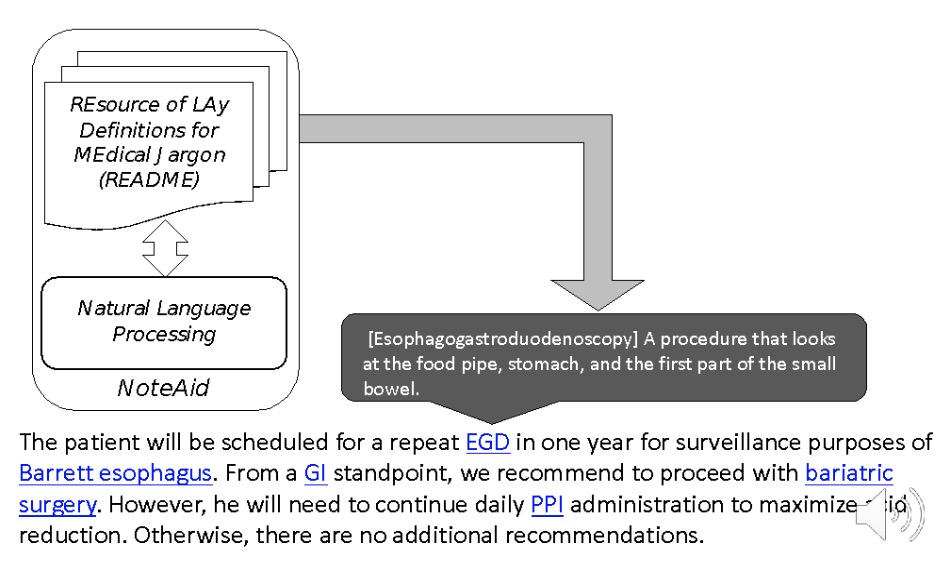}
    \caption{A visualization of the NoteAid pipeline, where NLP tools first identify jargon that may be challenging for patients to understand. The lay definitions corresponding to these jargon terms are then retrieved from relevant dictionaries and presented to the patients, enhancing their comprehension and engagement with their health information. 
    }
    \label{fig:noteaid}
\end{figure}

Throughout the extensive history of natural language processing (NLP), enhancing individuals' ability to read and comprehend complex texts has been a significant research goal that remains only partially achieved~\cite{zeng2020survey}.
Among the various tasks in this field, improving the comprehension of medical electronic health records (EHRs) stands out due to the complexity and specificity of medical terminology~\cite{nutbeam2023artificial}. 
Efforts to enhance the comprehension of EHRs can not only advance NLP's goal of aiding the understanding of complex texts but also hold substantial social significance by increasing efficiency and reducing errors for both patients and healthcare professionals~\cite{nanna2009health,HomeHealthCare, nurseNavigators, baldry1986giving, schillinger2009effects}.

However, despite advancements in EHRs management~\cite{walker2019opennotes}, one significant barrier persists in the form of medical jargon
~\footnote{A ``medical jargon term'' is the specialized language healthcare professionals use that can be complex for non-medical individuals. A ``lay definition'' translates this jargon into accessible language for the general public, aiming to bridge the understanding gap and enhance the communication of health-related information.}
in EHRs, impeding patient understanding and self-care~\cite{kujala2022patients,choudhry2016readability,khasawneh2022effect}. 
As shown in Figure~\ref{fig:noteaid}, tools like NoteAid~\cite{chen2018natural, kwon2022medjex}, which employs NLP to demystify complex medical terms, have been instrumental in bridging the communication gap between healthcare professionals and patients~\cite{lalor2018comprehenotes, lalor2021evaluating, lalor2023evaluating}. 
However, lay language dictionary resources like the Consumer Health Vocabulary (CHV)~\cite{zeng2006exploring, he2017enriching, ibrahim2020enriching} are limited in scale, posing a challenge to NoteAid. For instance, only a fraction (about 4\%) of the medical terms in NoteAid have been annotated with lay definitions, highlighting the need for a more scalable solution to address this knowledge gap effectively.

Addressing this issue requires shifting our focus to online health education materials such as Unified Medical Language System (UMLS) ~\cite{bodenreider2004unified}, MedlinePlus~\cite{patrias2007citing}, Wikipedia, and Google. 
However, as indicated in Table~\ref{tab:definitions}, these resources often present too difficult information for the average patient to understand. For example, these resources' average readability measured by the Flesch Kincaid Grade Level~\cite{solnyshkina2017evaluating} was post-secondary or higher education, while the average readability of a US adult was 7-8th grade level~\cite{doak1996teaching, doak1998improving, eltorai2014readability}.
To bridge this gap, we have engaged medical experts to meticulously curate lay definitions for jargon terms found in NoteAid-MedJEx~\cite{kwon2022medjex}, targeting a comprehension level suitable for individuals with a 7th to 8th-grade education.
Each jargon term has been redefined across various contexts, ensuring their applicability in diverse clinical scenarios. 
This effort led to our creation of the REsource of lAy Definitions for MEdical jargon (README) dataset, an expansive resource containing over 51,623 (medical jargon term, lay definition) pairs.
The README dataset comprises an impressive 308,242 data points, each consisting of a clinical note context, a medical jargon term, and its corresponding lay definition. Thus, the dataset significantly enhances the accessibility and comprehensibility of medical information for patients.

Yet, the critical aspect of generating lay definitions remains largely unexplored. 
As patients gain more access to their EHRs, the demand for lay definition resources is escalating. Despite our efforts to expand them, they are inevitably destined to surpass the capacity of current expert-annotated resources. 
Moreover, the dynamic nature of "jargon" based on individual and context makes pre-annotated expert resources less adaptable to real-life scenarios. 
The model-driven automatic generation of lay definitions from medical jargon emerges as a viable solution. 
Recent research highlighted ChatGPT's potential in its integration with the field of medicine~\cite{brown2020language,openai2023gpt4,yang2023performance}, including generating human-readable definitions for biomedical terms~\cite{remy2023automatic}. 
Nonetheless, our evaluation of open-source models (refer to Figure~\ref{fig:one_shot_performance}) shows a significant performance degradation compared to ChatGPT.
Using open-source large language models like Llama2~\cite{touvron2023llama} and small language models such as GPT-2~\cite{radford2019language} is crucial because proprietary LLMs accessed via third-party APIs may not always be feasible, especially in fields like healthcare with strict privacy requirements and economic constraints.
Open-source models offer the necessary privacy, while smaller models provide economic and infrastructural benefits, addressing distinct concerns about effectively deploying NLP tools in healthcare scenarios.

To bridge this gap, we aim to train an in-house system using open-source models for automatic lay definition generation to provide reliable lay definitions for jargon in patient education tools like NoteAid.
Inspired by research on Retrieval-Augmented Generation (RAG) in general and medical domains~\cite{lewis2020retrieval,asai2023self,xiong2024benchmarking,wang2024jmlr,guo2024retrieval}, we designed to use external resources to overcome the limitations of these open-source models in medical knowledge~\cite{sung2021can, yao2022extracting, yao2023context,chen2023meditron}.
We are positioning automatic lay definition generation as a form of text simplification, where language models are prompted to generate context-aware, jargon-specific, and layperson-friendly definitions based on general definitions retrieved from external knowledge resources. 
Specifically, in this work, we use the UMLS to retrieve general definitions of jargon terms and construct a dataset upon the README that includes context, jargon terms, general definitions, and lay definitions.

\begin{figure*}[!ht]
    \centering
    \includegraphics[width=\linewidth]{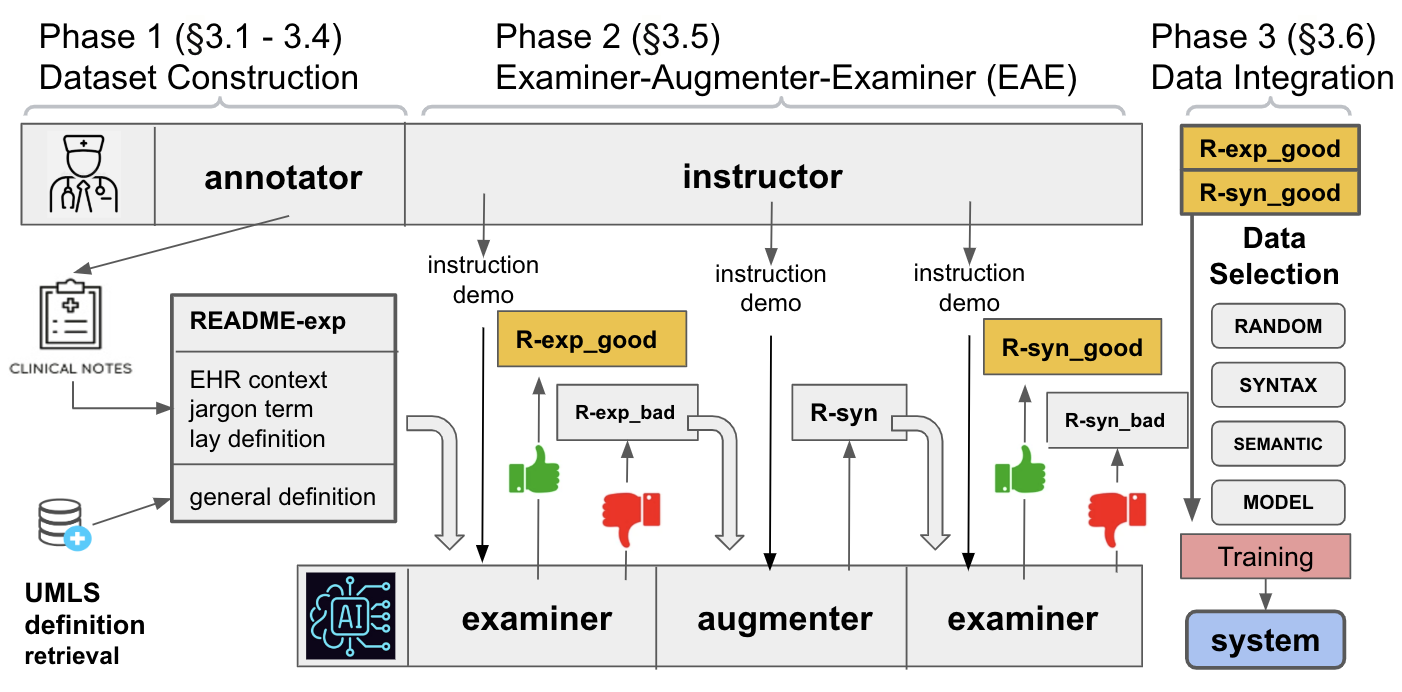}
    \caption{Our Data-Centric NLP pipeline, comprising 
    the Examiner-Augmenter-Examiner (EAE) framework and different data selection methods.
    EAE shows how humans (physicians) and AI (LLM, e.g. ChatGPT) cooperate to make a high-quality README dataset.
    We collect general definitions for every jargon term from external knowledge resources such as UMLS. ``R’' is ``README’'. 
    ``exp’' is ``expert annotation version’', ``syn’' is ``AI synthetic version’'. ``instruction’' and ``demo’' (examples for in-context learning) are combined into the prompt for LLM.
    In the pipeline, the human duties at different stages are annotator (labeling the initial dataset) and instructor (providing suitable prompts to guide AI at every stage). 
    The AI duties at different stages are examiner (filter high-quality data) and augmenter (improve the quality of low-quality data).
    Appendix Table \ref{tab:dataset statistics} describes the number of different versions of the README dataset in each step.
    After we get two high-quality datasets, R-exp\_good and R-syn\_good, from the EAE pipeline, we then deploy 4 different data selection strategies to combine high-quality expert-annotated data R-exp\_good and high-quality AI-synthetic data R-syn\_bad for in-house system training.
    }
    \label{fig:human-ai-loop}
\end{figure*}

To improve the initial README dataset's data quality for model training, we developed a data-focused process called Examiner-Augmenter-Examiner (EAE), as illustrated in Figure \ref{fig:human-ai-loop}.
Drawing inspiration from the human-in-the-loop concept~\cite{monarch2021human}, we employed human experts to guide AI in both the examiner and augmenter stages. The examiner filters high-quality training data, which may come from expert annotations before the augmenter stage or from AI-generated content after. The augmenter generates potentially high-quality synthesized data to increase data points. In the end of EAE, Human annotators review filtered data to ensure its quality.
After obtaining high-quality expert-annotated and AI-synthesized datasets, we employed the AI-synthesized dataset to augment the expert-annotated dataset, aiming to explore the effectiveness of AI-synthesized data in training.
We implemented a range of heuristic data selection strategies to integrate AI synthetic data, allowing us to incorporate suitable data points into our training process.

\noindent{\textbf{In summary, our contributions are as follows:}}
\begin{itemize}   
[leftmargin=.1in,topsep=0.1pt]
\setlength\itemsep{-0.5em}
\vspace{-0.2em}
    \item Introduced a new task of automatically generating lay definitions for medical jargon.
    We created a substantial expert-annotated dataset of 308,242 data points that can be used directly as a detailed lay-language dictionary for patient education tools such as NoteAid and as training data for this new task.
    \item Developed a robust, data-centric pipeline that effectively integrates data filtering, augmentation, and the selection of synthetic data. This approach enhances the quality of README datasets, merging the strengths of AI with human expertise to achieve optimal results.
    \item Our extensive automatic and human evaluations reveal that when trained with high-quality data, open-source, mobile-friendly small models can achieve or even exceed the performance of cutting-edge closed-source large language models, such as ChatGPT.

\end{itemize}

\section{Problem Statement}
\label{Sec:problem_statement}
Consider a dataset \(D = \{X, Y, Z_{+}\}\) comprising \(t\) EHRs, where \(X = \{x^1, x^2, \ldots, x^t\}\) represents the contexts of these EHRs, \(Y = \{y^1, y^2, \ldots, y^t\}\) denotes the corresponding jargon terms, and \(Z_{+} = \{z_{+}^1, z_{+}^2, \ldots, z_{+}^t\}\) are the ground truth expert lay definitions. Each EHR context \(x^i\) is a sequence of \(n\) tokens, expressed as \(x^i = \{x_1^{i}, x_2^{i}, \ldots, x_n^{i}\}\), and each lay definition \(z_{+}^i\) consists of \(m\) tokens, given by \(z_{+}^i = \{z_{+,1}^{i}, z_{+,2}^{i}, \ldots, z_{+,m}^{i}\}\). 
The README lay definition generation task \(T\) aims to train a reference model \(M_{ref}\) such that \(M_{ref}(z_{+}^i \mid x^i, y^i)\) is optimized. The standard approach for fine-tuning \(M_{ref}\) on \(T\) involves using the cross-entropy loss \(\ell_{ce}(z_{+}^i, M_{ref}(x^i, y^i))\) over the dataset \(D\).
To enhance the training of \(M_{ref}\), we introduce an additional set of general definitions \(Z_{-} = \{z_{-}^1, z_{-}^2, \ldots, z_{-}^t\}\), where each \(z_{-}^i\) corresponds to the general definition of the jargon term \(y^i\), generated using openly available data sources (UMLS) or GPT-3.5-turbo. Our proposed EAE pipeline is designed to acquire high-quality general definition data \(Z_{-}\), culminating in the augmented dataset \(D_{simp} = \{X, Y, Z_{+}, Z_{-}\}\).
The README lay definition generation task T is then formalized as a text simplification task, where \(M_{ref}\) is trained to produce \(Z_{+}\) based on \(X, Y,\) and \(Z_{-}\). This process utilizes a selected subset \(D_{SEL} \subseteq D_{simp}\), chosen according to one of the selection criteria: RANDOM, SYNTAX, SEMANTIC, or MODEL.

\section{Method}

\subsection{README Data Collection}
\label{Sec:data_collection}
The dataset source is a collection of publicly available deidentified EHR notes from hospitals affiliated with an anonymized institution. 
Herein, 18,178 sentences were randomly sampled, and domain experts then annotated the sentences for medical jargon and corresponding lay definition.

\subsection{Lay Definition Annotation} 
\label{Sec:data_annotation}

Domain-experts read each sentence and identified as medical jargon terms that would be considered difficult to comprehend for anyone no greater than a 7th-grade education~\footnote{The rule of thumb is that if a term has a lay definition comprehensible to a 4-7th grader as judged by FKGL~\cite{solnyshkina2017evaluating}, this term is included as a jargon term.}.
Overall, 51,623 unique (medical jargon term, lay definition) pairs with 308,242 mentions~\footnote{So there are 308,242 data points with (EHR context, medical jargon term, lay definition) format.} in the EHR repository have been annotated by complying with the annotation guidelines presented in Appendix~\ref{apx:annotation_guideline}.

\subsection{General Definition Retrieval}
\label{Sec:gen_def_retrival}

We then employed the Scispacy library~\footnote{We used the Scispacy \textit{en\_core\_sci\_lg} model to obtain the data. \url{https://github.com/allenai/scispacy}} to retrieve the corresponding UMLS definitions~\cite{lindberg1993unified} of these annotated medical jargon terms as the general definitions. 
We follow the retrieval and preprocessing steps in Appendix~\ref{apx:umls_def} to filter the valid general definitions for the README dataset~\footnote{We discuss the concept ambiguity issue in Appendix~\ref{apx:concept_ambiguity}}. 
This preliminary cleaning results in 308,242 data points in the README-exp (e.g., expert annotated) dataset, each consisting of a clinical note context, a medical jargon term, a corresponding lay definition, and a corresponding general definition, as shown in Figure~\ref{fig:human-ai-loop}.

\subsection{Examiner-Augmenter-Examiner (EAE)}
\label{Sec:data-centric}

\paragraph{Examiner (expert-annotated data)}
\label{Sec:examiner_expert}
Initially, basic data cleaning, as outlined in Appendix \ref{apx:umls_def}, was applied. 
To enhance this, we employed GPT-3.5-turbo, using a few-shot learning approach with seven examples - four demonstrating acceptable data points and three showing unacceptable ones. 
These prompts served as the `Human' element in our Human-AI-in-the-loop model, as depicted in Figure \ref{fig:human-ai-loop} and detailed in Algorithm~\ref{algo:data_cleaning}. 
The prompts are detailed in Table 9. 
We choose GPT-3.5-turbo here because our evaluation (Section~\ref{Sec:results} and Appendix~\ref{apx:soft_hard_correlation}) shows that the definitions it generates for medical terms can reach a human-satisfying level.
Post-cleaning~\footnote{More GPT Running Details are in Appendix~\ref{apx:gpt_run_details}}, approximately 39\% of UMLS general definitions were deemed suitable by the Examiner (e.g., GPT-3.5-turbo). 
The suitable 113,659 data points were archived in R-exp\_good, while the unsuitable 177,140 ones were stored in R-exp\_bad.

\paragraph{Augmenter}
Given the low yield of usable UMLS definitions, we employed GPT-3.5-turbo to augment our dataset.
The augmentation process, a critical part of our Data-centric pipeline, involved the system prompt: ``\emph{Generate a general definition of the term.}'' 
This step, accompanied by two examples (Table 9), aimed to create correct general definitions but may not be suitable for laypeople (similar to the UMLS definitions). 
The outcome of this process was R-syn (e.g., AI synthetic), containing 171,831 newly generated definitions.

\paragraph{Examiner (AI-synthetic data)}
The ChatGPT-generated definitions underwent a second cleaning round using the same methodology as in Section~\ref{Sec:examiner_expert} Examiner (expert-annotated data). 
Here, approximately 56\% of the ChatGPT definitions were found suitable for model training, with the remaining being either contextually inappropriate or incompatible with the expert-provided lay definitions. The final tally was 96,668 `good' and 75,175 `bad' general definitions, stored in R-syn\_good and R-syn\_bad, respectively.
We discuss the EAE pipeline efficacy in Appendix~\ref{apx:eae_efficacy}

\paragraph{Qaulty Checking} After the end of the EAE, we sampled 500 medical jargon terms each from the R-exp\_good and R-syn\_good datasets. 
We conducted human verification~\footnote{The details about Data Quality Checking and Train/Eval/Test Split can be found in Appendix~\ref{apx:soft_hard_correlation}} on corresponding data points for these 1000 jargon terms. 
We obtained a high human agreement for both the R-exp\_good and R-syn\_good datasets' quality.
After correcting individual invalid data, we used this portion of the data as evaluation and test data (in a 1:1 ratio). 
We used the remaining R-exp\_good and R-syn\_good datasets as training data, ensuring that these 1,000 medical jargon terms would not appear in the training data. 
Table \ref{tab:dataset statistics} shows the overall dataset statistics of all the README versions across the pipeline.

\subsection{Integration of Synthetic and Expert Data}
\label{Sec:data-selection}
We adopted four distinct sampling strategies to integrate the AI-synthetic training data (e.g., R-syn\_good) into the expert-annotated training data (e.g., R-exp\_good):

\begin{itemize}
[leftmargin=.1in,topsep=0.1pt]
\setlength\itemsep{-0.5em}
\vspace{-0.2em}
    \item \textbf{RANDOM:} This approach randomly selected N entries from the R-syn\_good dataset. This is the baseline for our subsequent three heuristic methods.
    \item \textbf{SYNTAX:} For the syntax-based sampling approach, the ROUGE\_L metrics F1 score in Section~\ref{Sec:evaluation} was used as a key evaluative tool. ROUGE\_L focuses on the longest common subsequence, which measures the longest string of words that occurs in both the predicted and reference texts. 
    By using this metric, we could rank the synthetic definitions according to their syntactic closeness to the human-written definitions, which helped us select samples that would potentially be more understandable and natural-sounding.
    \item \textbf{SEMANTIC:} For semantic-based sampling, we utilized SentenceTransformers~\footnote{We used default model \textit{all-MiniLM-L6-v2} in \url{https://github.com/UKPLab/sentence-transformers}.}. 
    Renowned for its text semantic analysis efficiency, this model enabled us to measure the semantic similarity between lay definitions in R-exp\_good and R-syn\_good datasets. We ranked synthetic data based on these scores, considering higher scores as indicative of greater semantic closeness to expert annotations.
    \item \textbf{MODEL:} In model-based sampling, 
    we used models initially trained on the R-exp\_good dataset to generate definitions for the R-syn\_good dataset. We employed the ROUGE\_L F1 score to evaluate the alignment between model-generated and actual R-syn\_good lay definitions. 
    This technique aids in mitigating training challenges associated with data heterogeneity. It enriches the dataset with examples that enhance the model's convergence towards the desired distribution (e.g., expert-annotated lay definitions).
\end{itemize}

\section{Experiments}
\label{Sec:experiments}
\subsection{Automatic Evaluation Metrics}
\label{Sec:evaluation}
We evaluate the efficacy of our model in producing lay definitions by contrasting them with the ground-truth reference lay definitions, utilizing the ROUGE~\cite{lin2004rouge} and METEOR~\cite{banarjee2005} metrics. 
However, they are based on the exact word overlap and therefore provide insight into the informativeness of the generated lay definitions but do not necessarily reflect their factual accuracy~\cite{maynez2020faithfulness}. 
We also employ Scispacy to extract medical concepts from the model-generated and the reference lay definitions. 
We then compute the F1 Scores for these concept lists, referred to as UMLS-F1, to specifically measure the factuality of the generated content~\footnote{The details about UMLS-F1 can be found in Appendix~\ref{apx:umlsf1}}.

\subsection{Experimental Setting}
\label{Sec:experimental_setting}
We use the following base models: GPT-2, DistilGPT2, BioGPT, and Llama2 in our experiments~\footnote{More Experimental Setting details are in Appendix~\ref{apx:exp_setting}.}. 
We use the following symbols:
\begin{enumerate}[topsep=0.5pt,itemsep=0.2ex,partopsep=0.2ex,parsep=.20ex]
    \item jargon2lay(J2L): Directly generates a lay definition for a given jargon term.
    \item jargon+context2lay(J+C2L): Generates a lay definition for a given jargon term based on the context information from clinical documents.
    \item jargon+gen2lay(J+G2L): Generates a lay definition for a given jargon term based on the general definition provided by UMLS.
    \item jargon+context+gen2lay(J+C+G2L): Generates a lay definition for a given jargon term based on both the context information from clinical documents and the general definition provided by UMLS.
\end{enumerate}

We use one jargon, 'EDG', as an example and show the input prompt of different settings J2L, J+C2L, J+G2L, J+C+G2L~\footnote{We use "context" and "C" interchangeably to refer to the EHR context in which the jargon occurs. Similarly, we use "general definition" and "G" interchangeably to refer to the general definition retrieved from UMLS.} on Table~\ref{different_settings_prompt}.

Our experiments were divided into five distinct parts.
In \textbf{Set-1}, we aimed to evaluate the performance gap on the J2L task between open-source base models and GPT3.5/4 models. 
We do one-shot prompting for Llama2~\footnote{Because smaller open source language models DistilGPT2/GPT2/BioGPT do not have the ability to complete instruction following under zero-shot or few-shot settings, we only compare the results of Llama2 as a representative of open source language models with GPT3.5/4 to see the gap.}, GPT-3.5-turbo, and GPT-4 with prompts in Table~\ref{one_shot_prompt}. 
\textbf{Set-2} explored the varying data quality across different versions within our EAE pipeline, where we fine-tuned the base models on J+G2L task.
\textbf{Set-3} focused on evaluating the effects of various data selection strategies on data augmentation outcomes. 
To do this, we ranked AI-synthetic data (R-syn\_good) using different methods in Section \ref{Sec:data-selection}, selecting either the top N entries with the highest scores (e.g., `SEMANTIC' in Table~\ref{table:data_selection_results}) or the bottom N entries with the lowest scores (e.g., `SEMANTIC\_r'). 
These selections were then evenly combined with expert-annotated data (R-exp\_good) at a one-to-one ratio. 
Following this, we fine-tuned the base models using these diverse, mixed datasets to determine the impact of each selection method on model performance.
In \textbf{Set-4}, we investigated the effects of incorporating different types of information (EHR context and UMLS-retrieved general definition) into the model inputs. 
In \textbf{Set-5}, we fine-tuned models of varying sizes (ranging from DistilGPT2-88M to Llama2-7B) and compared the outcomes with GPT-3.5-turbo using best settings learned from previous Set-2, Set-3, and Set-4. More experiment designs and results can be found in Appendix~\ref{apx:exp_design_result}

\subsection{Results}
\label{Sec:results}

\begin{figure}
    \centering
    \includegraphics[width=\linewidth]{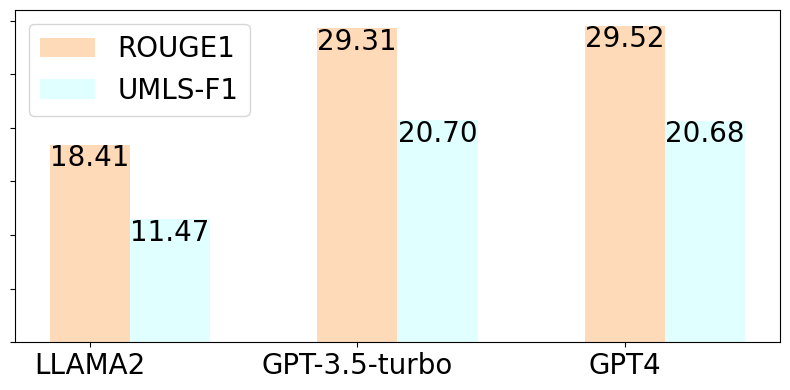}
    \caption{One-shot performances on jargon2lay.}
    \label{fig:one_shot_performance}
\end{figure}

We start with our experiments with some base models' performance on the J2L dataset with \textbf{Set-1}.
In Appendix \ref{apx:soft_hard_correlation}, we showed a high level of human evaluators' agreement for the definitions generated by GPT-3.5-turbo, which underlines its efficacy in crafting human-readable explanations for biomedical concepts.
Consequently, GPT-3.5-turbo serves as a strong baseline for quality in README lay definition generation, indicative of a standard that meets human satisfaction.
Despite this, our analysis of Llama2, illustrated in Figure~\ref{fig:one_shot_performance}, reveals a significant performance gap. This discrepancy underscores the critical necessity for enhancing the capabilities of Llama2 and other open-source models to achieve high-quality output in lay definition generation tasks, thereby improving patient education together with systems like NoteAid. 
Additionally, we observed that GPT-3.5-turbo and GPT4~\footnote{All GPT experiments in our paper were conducted using Microsoft Azure, which can be used HIPAA-compliantly, ensuring the ethical handling of sensitive data.} exhibited comparable proficiency in this task.

\subsubsection{Effectiveness of EAE pipeline}
\begin{table}
\centering
\scalebox{0.6}{
\begin{tabular}{c|ccccc|c}
\hline

\hline
J+G2L  & \small{ROUGE1} &  \small{ROUGE2} & \small{ROUGEL} & \small{METEOR} & \small{UMLS-F1} & \small{Rank}
\\

\hline

R-exp & 23.94 & 8.95 & 22.79 & 17.75 & 12.83 & 4
\\
R-exp\_good & 26.99 & 10.76 & 25.57 & 21.19 & 16.88 & 2
\\
R-exp\_bad & 22.25 & 7.46 & 21.09 & 17.18 & 11.72 & 5
\\
R-syn\_good & 25.71 & 9.75 & 24.41 & 20.23 & 16.01 & 3
\\
R-exp+syn\_good & \color{red}{29.82} & \color{red}{13.14} & \color{red}{28.47} & \color{red}{24.42} & \color{red}{20.09} & 1
\\

\hline

\end{tabular}
}
\caption{Various README versions data performance.} 
\label{table:EAE_results}
\end{table}

In \textbf{Set-2}, we focus on the efficacy of different data versions when finetuning with the GPT-2 model. 
The results, as reflected in Table~\ref{table:EAE_results}, indicate that high-quality expert data (R-exp\_good) demonstrates clear superiority over unexamined expert data (R-exp), emphasizing the crucial role of ECE-examiner in enhancing data quality. 
Furthermore, high-quality synthetic data (R-syn\_good) outperforms the unexamined expert data (R-exp), underscoring the significant value of ECE-augmenter. 
Notably, the combination of R-exp\_good and R-syn\_good shows improved performance over R-exp\_good alone, suggesting that including synthetic data is beneficial.
This composite approach of R-exp\_good+R-syn\_good leading the rank underscores the efficacy of our EAE pipeline.

\subsubsection{Expert and Synthetic Data Integration}

\begin{table}
\centering
\scalebox{0.62}{
\begin{tabular}{c|cccc}
\hline

\hline
\small{ROUGE1} &  \small{J2L} & \small{J+C2L} & \small{J+G2L} & \small{J+C+G2L}
\\

\hline
RANDOM & 19.03 & 19.65 & 26.21 & 26.97
\\
SYNTAX(R) & 19.82(\textcolor{red}{-2.28}) & 20.65(\textcolor{red}{-2.78}) & 27.56(\textcolor{red}{-8.36}) & 28.33(\textcolor{red}{-8.84})
\\
SEMANTIC(R) & 19.74(-1.11) & 19.94(-1.39) & 26.42(-0.92) & 27.19(-1.96)
\\
MODEL(R) & 20.18(-0.8) & 20.46(-1.88) & 27.9(-2.37) & 28.65(-4.22) 
\\
\hline

\small{UMLS-f} &  \small{J2L} & \small{J+C2L} & \small{J+G2L} & \small{J+C+G2L}
\\

\hline
RANDOM & 8.05 & 9.09 & 15.78 & 16.72
\\
SYNTAX(R) & 8.53(-0.11) & 9.99(\textcolor{red}{-1.00}) & 17.76(\textcolor{red}{-5.11}) & 17.42(\textcolor{red}{-5.54})
\\
SEMANTIC(R) & 8.98(-0.38) & 9.07(-0.69) & 15.93(-0.02) & 16.74(-0.93)
\\
MODEL(R) & 9.18(\textcolor{red}{-0.63}) & 9.4(-0.78) & 18.34(-2.5) & 18.45(-2.55)
\\

\hline

\end{tabular}
}
\caption{Different data selection methods performance. The values in parentheses represent the difference between the corresponding X\_r (bottom N) and X (top N), i.e., X\_r - X. The smaller this value, the more it indicates that the X method can select higher quality synthetic data for data augmentation; conversely, the closer this value is to 0, the more it suggests that the X method cannot identify the higher-quality synthetic data for training.
} 
\label{table:data_selection_results}
\end{table}

In \textbf{Set-3}, we explored the effects of various data selection strategies on data augmentation outcomes. 
We found the results of SEMANTIC are closer to RANDOM than SYNTAX and MODEL, and SEMANTIC\_r - SEMANTIC is close to 0.
However, significant differences can be observed in SYNTAX and SYNTAX\_r, MODEL and MODEL\_r.
Table~\ref{table:data_selection_results} highlights two main findings: 
firstly, data selection is crucial, as all methods have better performance with higher-ranked data (e.g., SYNTAX, SEMANTIC, MODEL) over lower (e.g., SYNTAX\_r, SEMANTIC\_r, MODEL\_r). 
All three methods, SYNTAX, SEMANTIC, and MODEL, have better results than the RANDOM baseline.
Secondly, SYNTAX and MODEL are more effective in selecting higher-quality synthetic data for data augmentation than SEMANTIC and RANDOM. 
We selected SYNTAX as the default data augmentation method for subsequent experiments.


\subsubsection{Retrieval-augmented Generation}
\begin{table}[H]
\centering
\scalebox{0.62}{
\begin{tabular}{c|ccccc}
\hline

\hline
& \small{ROUGE1} &  \small{ROUGE2} & \small{ROUGEL} & \small{METEOR} & \small{UMLS-F1}
\\

& \multicolumn{5}{c}{Without data augmentation (R-exp\_good)}
\\
\hline

J2L & 19.47 & 6.06 &	18.53 &	14.74 & 8.76
\\
J+C2L & 19.40 & 6.40 & 18.38 & 15.12 & 9.24
\\
J+G2L & 26.99 & 10.76 & 25.57 & 21.19 & 16.88
\\
J+C+G2L & \textcolor{red}{27.58} & \textcolor{red}{11.31} & \textcolor{red}{26.35} & \textcolor{red}{21.72} & \textcolor{red}{17.12}
\\

\hline
& \multicolumn{5}{c}{Data augmentation (R-exp+syn\_good with SYNTAX)}
\\
\hline

J2L & 21.98 & 7.42 &	20.88 & 16.98 & 10.95
\\
J+C2L & 22.13 & 7.90 & 21.04 & 17.54 & 10.71
\\
J+G2L & 29.82 & 13.14 & 28.47 & 24.42 & 20.09
\\
J+C+G2L & \textcolor{red}{29.89} & \textcolor{red}{13.48} & \textcolor{red}{28.49} & \textcolor{red}{24.65} & \textcolor{red}{20.27}
\\

\hline

\end{tabular}
}
\caption{Efficacy of incorporating EHR context and general definition in input data.
The retrieved general definitions significantly aid the overall performance of the model (ROUGE and METEOR) and also reduce hallucinations (UMLS factuality score).
} 
\label{table:jargon+gen+context}
\end{table}

\textbf{Set-4} results underscore the significant improvement in model performance when input data is enriched with UMLS-retrieved general definitions. 
As illustrated in Table~\ref{table:jargon+gen+context}, regardless of whether we utilize only expert-annotated data or data augmentation with AI-synthetic data, including general definitions consistently enhances effectiveness. 
This finding confirms the value of RAG with the general definition in the lay definition generation task. 
Meanwhile, adding EHR context to the input data yields a moderate impact on model performance.

\subsubsection{Model Performances Against ChatGPT}

\begin{figure}[H]
    \centering
    \includegraphics[width=\linewidth]{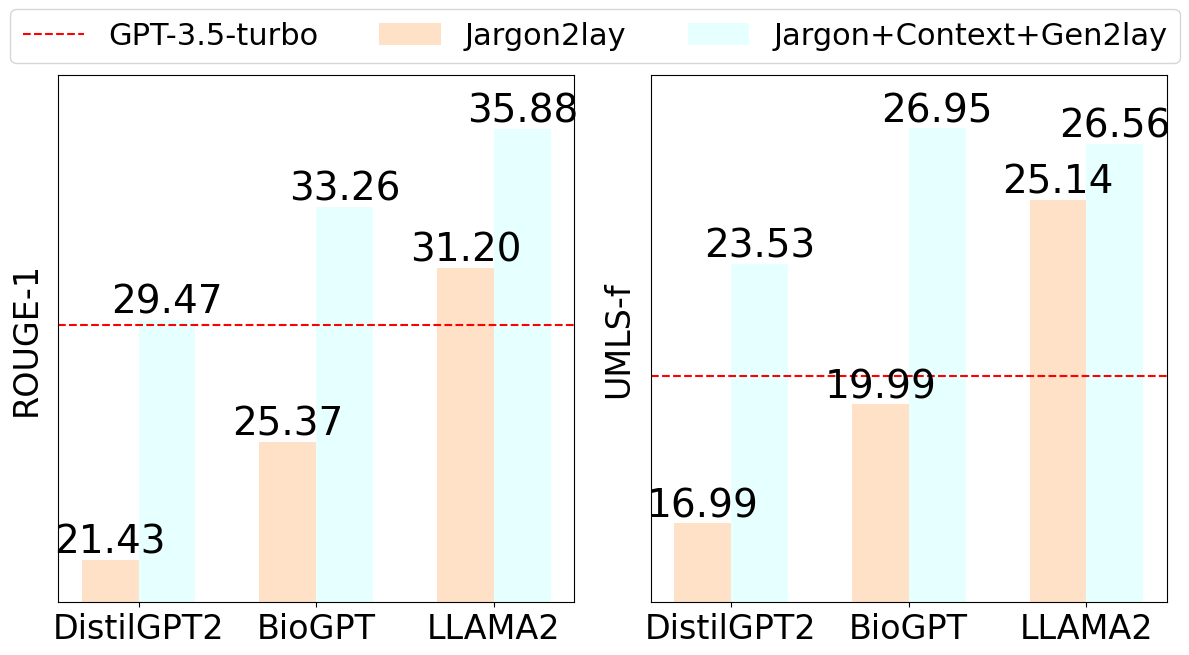}
    \caption{Comparative performance analysis of DistilGPT2, BioGPT, and LLAMA2 against GPT-3.5-turbo.}
    \label{fig:gpt2_biogpt_llama2_biollama2}
\end{figure}

In \textbf{Set-5}, we observed that LLAMA2-7B's ROUGE-1 and UMLS-F1 metrics surpassed GPT-3.5-turbo in the J2L task post-training. 
For the J+G2L setting, DistilGPT2-88M demonstrated equivalent results to GPT-3.5-turbo, while BioGPT's performance exceeded it, and LLAMA2-7B significantly outperformed the GPT-3.5-turbo. 
These findings, as depicted in Figure~\ref{fig:gpt2_biogpt_llama2_biollama2}, emphasize the effectiveness of open-source, mobile-adapted small models when appropriately fine-tuned with high-quality datasets, offering a promising avenue for deploying lightweight yet powerful NLP tools in mobile healthcare applications to help patient education.

\section{Human Evaluation}
\label{Sec:human_eval}
\subsection{Human Evaluation settings}
Our human evaluation was conducted by 5 human evaluators~\footnote{Since the generated lay definition is provided to lay people, five people without medical background were found to conduct the human evaluation here.}. 
We randomly selected 50 pairs of (jargon, generated lay definitions) from the test dataset for this human evaluation. 
The task for evaluators was to reference the expert definitions and choose a binary preference among the following four groups of definitions: 
1) DistilGPT2-J2L vs. GPT-3.5-turbo, 
2) DistilGPT2-J+C+G2L vs. GPT-3.5-turbo,
3) LLAMA2-J2L vs. GPT-3.5-turbo, 
4) LLAMA2-J+C+G2L vs. GPT-3.5-turbo. 
After we get judgments from multiple people per instance, we do not aggregate their labels before calculating the win rate but count them individually.

\subsection{Human Evaluation Results}

\begin{figure}[H]
    \centering
    \includegraphics[width=\linewidth]{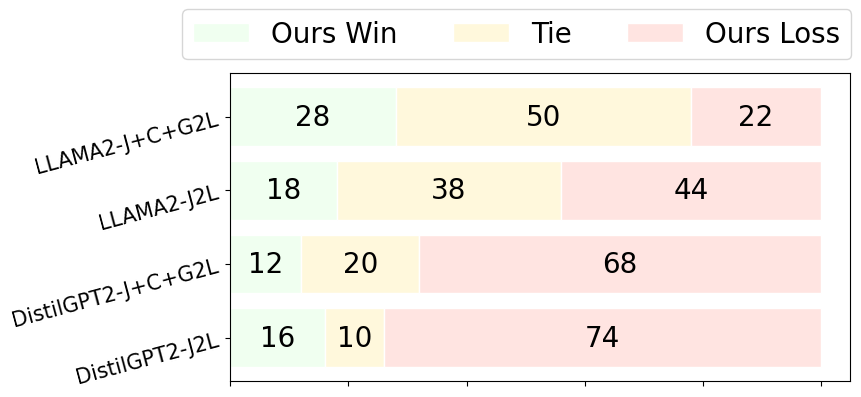}
    \caption{Human evaluation results (win rate \%).}
    \label{fig:human_eval_results}
\end{figure}

As shown in Figure~\ref{fig:human_eval_results}, although the result of adding EHR context and general definition (DistilGPT2-J+C+G2L) is better than DistilGPT2-J2L, the win rate of the two DistilGPT2 models' output is still significantly behind the result of GPT-3.5-turbo. For LLAMA-2, generating lay definitions directly from jargon is still not as good as GPT-3.5-turbo, but adding context and general definitions is of great help. Human evaluators prefer LLAMA2-J+C+G2L more than GPT-3.5-turbo.
There are some inconsistencies between the results of human evaluation and automatic evaluation. We further interviewed our medical experts about the reasons for our system win or loss cases and concluded the following conclusions to help future improvements:
1. While all our in-house systems perform satisfactorily, GPT-3.5-turbo stands out for its flexibility and user-friendliness. It excels at elaborating complex medical terms, offering detailed explanations and practical examples to improve comprehension.
2. Recent advancements~\cite{cai2023paniniqa, zhang2023ehrtutor} reveal ChatGPT's role in enhancing patient education through interactive formats like NoteAid-interactive~\cite{NoteAid-interactive}. 
It enables patients to actively ask questions and seek clarifications while the AI tailors responses to aid their understanding. 
This interactive approach, absent in traditional dictionary-style definitions like our README dataset, calls for next-step model distillation work or further refinement in aligning the in-house system's outputs with patient preferences. 
3. Additionally, developing automatic metrics aligning closely with human evaluation is another critical next step.

\section{Related Work}
\label{Sec:related_work}

The evolution of patient-centric healthcare necessitates simplified patient access to medical information. Tools like NoteAid and MedJEx have initiated efforts to make EHR content more comprehensible \citep{chen2018natural, kwon2022medjex}. 
However, sentence-level text simplification efforts have been expanded to larger datasets that capture broader biomedical contexts \citep{jonnalagadda2010towards, guo2021automated, devaraj2021paragraph}.
Recent research has similarly focused on the development of datasets and methods for text simplification, primarily at the sentence level, with the expansion into datasets capturing broader contexts of biomedical abstracts \citep{cao2020expertise, lu2023napss, goldsack2022making, luo2022readability}. 
Notably, efforts such as the CELLS, PLABA, and AGCT datasets have contributed significantly to this domain, providing extensive resources for training models capable of translating scientific discourse into lay language \citep{guo2024retrieval, attal2023dataset,remy2023automatic}.
Our work diverges from these existing efforts by introducing the README dataset, an expansive collection explicitly designed for context-aware lay definitions, addressing the nuanced task of generating patient-friendly definitions directly from medical terms, filling a critical gap in patient education resources.

In line with advancing the quality of generated texts, we have embraced Retrieval-Augmented Generation (RAG) to mitigate common issues in natural language generation, such as "hallucinations" \citep{karpukhin2020dense, shuster2021retrieval}.
Two main categories of information retrieval methods have been used to augment the generation of biomedical natural language generation, definition-based and embedding-based retrieval techniques \citep{guo2024retrieval, alambo2022entity, moradi2018different, xiong2024benchmarking}. Our RAG belongs to the definition-based retrieval technique.

Finally, our contributions distinctly highlight the integration of a robust, data-centric Human-AI pipeline that improves data quality and the efficiency of models trained on the README dataset. 
This innovative pipeline leverages the Data-centric AI framework, navigating through phases of collection, labeling, preparation, reduction, and augmentation to build a dataset that is both expansive and representative \cite{zha2023data, ng2021data, whang2023data}. 
The process begins with meticulous data collection and expert labeling, ensuring a foundation of high-quality, domain-specific data (Section \ref{Sec:data_collection} - \ref{Sec:gen_def_retrival}). 
During the preparation phase, raw data undergoes rigorous cleaning and transformation, readying it for effective model training \cite{krishnan2019alphaclean}. 
The dataset is then enriched through strategic data augmentation techniques, incorporating verified quality AI-synthetic data to expand its scope and utility (Section \ref{Sec:data-centric}). 
Furthermore, data reduction strategies are employed to select the most suitable instances for integration, enhancing the dataset's overall effectiveness (Section \ref{Sec:data-selection}). 
Through these meticulous stages, the README dataset not only supports but significantly enhances the capabilities of smaller, open-source models, allowing them to match or even exceed the performance of larger proprietary models like ChatGPT in specific healthcare applications.

\section{Conclusions}
\label{Sec:conclusions}
Our study underscores the potential of NLP to democratize medical knowledge, enabling patient-centric care by simplifying complex medical terminology. 
Developing the README dataset and implementing a data-centric pipeline has improved dataset quality and expanded AI training possibilities. 
Our experiments show that small open-source models can match advanced close-source LLMs like ChatGPT with well-curated data.
README will be open to the community as an important lay dictionary for patient education.
We hope our work can help the innovative patient education community advance toward a future where all patients can easily understand their health information.

\section{Limitations and Ethical Considerations}
\label{sec:ethics}
This study provides valuable insights, but experimental results evaluated in humans demonstrate limitations of the current work and some future directions. First, better automatic evaluation metrics need to be explored to be closer to human evaluation results. Secondly, in this paper, we have only explored some heuristic data selection methods, and we need to explore more sophisticated methods in the future. In addition, the next step of the in-house system is to collect patient preferences for human alignment, which can help us generate a more user-friendly or customized lay definition. Also, we can use ChatGPT or LLAMA2-J-C-G2L to serve as the teacher and use DistilGPT2-based systems to serve as the students, performing distillation to improve the performance of the small models post current supervisor-fine-tuning on README. Finally, more interactive ways need to be considered in the future to make the in-house system more user-friendly and patient-centric.

Consider Privacy Implications, LLMs (especially third-party APIs like ChatGPT) may raise privacy concerns when conducting patient education, which may violate HIPAA regulations. In this study, we manually annotated lay definitions on publicly available MedJEx jargon terms and obtained general definitions from accessible UMLS. We also make AI-synthetic data to help training since synthetic data generation is an active field in the clinical domain especially to overcome privacy concerns~\cite{pereira2022secure,shafquat2022source,mishra2023synthetic}.
The trained in-house system can be deployed on the patient's mobile to avoid patient data leaving the local area, which can better protect the patient's privacy and security. Consider Biases, LLMs trained on large amounts of text data may inadvertently capture and reproduce biases present in the data. Therefore, an in-house system trained on our data (whether expert annotation or AI synthetic) may perpetuate incorrect information or provide inaccurate answers. Finally, although we used UMLS-based RAG to reduce hallucinations, LLMs may still generate factual errors when conducting patient education.

\section*{Acknowledgements}
We sincerely appreciate the tremendous efforts of the entire README annotation team throughout the expert annotation process of the README dataset. In particular, we would like to extend our heartfelt thanks to Weisong Liu, Harmon S. Jordan, David A. Levy, and Brian Corner for their valuable contributions and dedication to this project. Their expertise and hard work have been integral to the success of this endeavor.

\bibliography{emnlp}
\bibliographystyle{acl_natbib}


\appendix

\begin{table*}[h!]
\centering
\begin{tabular}{|l|p{12cm}|l|}
\hline
\textbf{Source} & \textbf{Definition} & \textbf{FKGL} \\
\hline
UMLS & An endoscopic procedure that visualizes the upper part of the gastrointestinal tract up to the duodenum. & 13.5 \\
\hline
MedlinePlus & Esophagogastroduodenoscopy (EGD) is a test to examine the lining of the esophagus, stomach, and first part of the small intestine (the duodenum). & 16.1 \\
\hline
Wikipedia & Esophagogastroduodenoscopy (EGD), also called by various other names, is a diagnostic endoscopic procedure that visualizes the upper part of the gastrointestinal tract down to the duodenum. It is considered a minimally invasive procedure since it does not require an incision into one of the major body cavities and does not require any significant recovery after the procedure (unless sedation or anesthesia has been used). & 20.9 \\
\hline
Google & An EGD is a procedure in which a thin scope with a light and camera at its tip is used to look inside the upper digestive tract – the esophagus, stomach, and first part of the small intestine, called the duodenum. It's also called an upper endoscopy, or an esophagogastroduodenoscopy. & 13.2 \\
\hline
README & [Esophagogastroduodenoscopy] A procedure that looks at the food pipe, stomach, and the first part of the small bowel. & 5.6 \\
\hline
\end{tabular}
\caption{Definitions of Esophagogastroduodenoscopy from various sources.}
\label{tab:definitions}
\end{table*}

\begin{table*}[h!]
\centering
\begin{tabular}{|l|p{9cm}|l|}
\hline
\textbf{README-version} & \textbf{Dataset Description} & \textbf{DataPoints} \\
\hline
README-exp & (ehr context, jargon, lay def, general definition) & 308,242 \\
\hline
README-exp & (jargon, lay def, general definition)& 51,623 \\
\hline
README-exp\_good & (ehr context, jargon, lay def, general definition) &  113,659 \\
\hline
README-exp\_good & (jargon, lay def, general definition) & 11,765 \\
\hline
README-exp\_bad & (ehr context, jargon, lay def, general definition) &  177,140 \\
\hline
README-exp\_bad & (jargon, lay def, general definition) & 39,856 \\
\hline
README-syn & (ehr context, jargon, lay def, general definition) &  177,140 \\
\hline
README-syn & (jargon, lay def, general definition) & 39,856 \\
\hline
README-syn\_good & (ehr context, jargon, lay def, general definition) &  96,668 \\
\hline
README-syn\_good & (jargon, lay def, general definition) & 96,668 \\
\hline
README-syn\_bad & (ehr context, jargon, lay def, general definition) &  75,157 \\
\hline
README-syn\_bad & (jargon, lay def, general definition) & 75,157 \\
\hline
\end{tabular}
\caption{The Dataset Statistics of Different README versions.}
\label{tab:dataset statistics}
\end{table*}

\section{Annotation Guideline}
\label{apx:annotation_guideline}
The dataset was annotated for medical jargon and lay definition by six domain experts from medicine, nursing, biostatistics, biochemistry, and biomedical literature curation~\footnote{The annotator agreement scores can be found in Appendix~\ref{apx:annotation_reliability}.}. 
Herein, the annotators applied the following rules for identifying what was jargon and how to write a suitable lay definition:

\noindent \textbf{Rule 1.} Medical terms that would \textbf{not be recognized by about 4 to 7th graders}, or that \textbf{have a different meaning in the medical context than in the lay context (homonym)} were labeled. 
For example:

\begin{itemize}
[leftmargin=.1in,topsep=0.1pt]
\setlength\itemsep{-0.5em}
\vspace{-0.2em}
    \item accommodate: When the eye changes focus from far to near.
    \item antagonize: A drug or substance that stops the action or effect of another substance.
    \item resident: A doctor who has finished medical school and is receiving more training.
    \item formed: Stool that is solid.
\end{itemize}

\noindent\textbf{Rule 2.} Terms that are not strictly medical, but are \textbf{frequently used in medicine}. For example:

\begin{itemize}
[leftmargin=.1in,topsep=0.1pt]
\setlength\itemsep{-0.5em}
\vspace{-0.2em}
    \item "aberrant", "acute", "ammonia", "tender", "intact", "negative", "evidence"
\end{itemize}
 
\noindent\textbf{Rule 3.} When jargon words are \textbf{commonly used together, or together they mean something distinct or are difficult to understand from the individual parts quickly} were labeled. For example:

\begin{itemize}
[leftmargin=.1in,topsep=0.1pt]
\setlength\itemsep{-0.5em}
\vspace{-0.2em}
    \item vascular surgery: Medical specialty that performs surgery on blood vessels.
    \item airway protection: Inserting a tube into the windpipe to keep it wide open and prevent vomit or other material from getting into the lungs.
    \item posterior capsule: The thin layer of tissue behind the lens of the eye. It can become cloudy and blur vision.
    \item right heart: The side of the heart that pumps blood from the body into the lungs.
    \item intracerebral hemorrhage: A stroke.
\end{itemize}

\noindent\textbf{Rule 4.} Terms whose \textbf{definitions are widely known} (e.g., by a 3rd grader) do NOT need to be labeled. For example:

\begin{itemize}
[leftmargin=.1in,topsep=0.1pt]
\setlength\itemsep{-0.5em}
\vspace{-0.2em}
    \item “muscle”, “heart”, “pain”, “rib”, “hospital”
\end{itemize}

\textbf{Rule 4.1} When in doubt, label the term. For example: 
\begin{itemize}
[leftmargin=.1in,topsep=0.1pt]
\setlength\itemsep{-0.5em}
\vspace{-0.2em}
    \item “colon”, “immune system”
\end{itemize}

\section{Evaluation of the Annotation}
\label{apx:annotation_reliability}
An observational study was performed to evaluate the annotators' reliability in identifying jargon and providing lay definitions, and assess the agreement of the dataset annotators with each other and laypeople.

\subsection{Data Collection and Setting}
For evaluation, twenty sentences were randomly selected from deidentified inpatient EHR notes in the EHR repository of one hospital affiliated with an anonymized institution.
Sentences consisting only of administrative data, sentences less than ten words long, and sentences substantially indistinguishable from another sentence were filtered out. 

Note that the annotators had never seen the sampled sentences. The twenty sentences were made up of 904 words in total. 
Common words were discarded so as not to inflate the calculated agreement. These consisted of all pronouns, conjunctions, prepositions, numerals, articles, contractions, months, punctuation, and the most common 25 verbs, nouns, adverbs, and adjectives. 
Terms occurring more than one time in a sentence were counted only once. Furthermore, multi-word terms were counted as single terms to ameliorate the double-counting issue. 
Two members of the research team determined multi-word terms by consensus. In this work, multi-word terms were defined as adjacent words that represented a distinct medical entity (examples: “PR interval”, “internal capsule”, “acute intermittent porphyria”), were commonly used together (examples: “hemodynamically stable”, “status post”, “past medical history”) and terms that were modified by a minor word (examples: “trace perihepatic fluid”, “mild mitral regurgitation”, “rare positive cells”, “deep pelvis”).
After applying these rules, 325 candidate medical jargon terms and their lay definition were utilized. 
The laypeople comprised 270 individuals recruited from Amazon Mechanical Turk (MTurk) \citep{aguinis2021mturk}.

\subsection{Annotation Reliability}
The results showed that there was good agreement among annotators (Fleiss’ kappa = 0.781). 
The annotators had high sensitivity (91.7\%) and specificity (88.2\%) in identifying jargon terms and providing suitable lay definitions as determined by the laypeople (the gold standard).

\subsection{Details about Jargons and Lay Definitions Statistics in README-exp}
\label{apx:stats_detail}

\subsection{GPT Running Details in Examiner step}
\label{apx:gpt_run_details}
To optimize computing resources, we streamlined our dataset by removing EHR context and focusing solely on unique data points with (medical jargon term, lay definition, general definition) format in this Examiner step~\footnote{This was done because, in many cases, we have the same jargon term, general definition from UMLS, and lay definition but different EHR contexts. We can reduce the amount of GPT running if we ignore the EHR context difference.}. This reduced our dataset from 308,242 to 51,623 data points. 
Upon processing these through Examiner, 11,765 were classified as `good' quality general definitions, and 39,856 as `bad'. 
We subsequently performed an SQL join operation, integrating the `good' and `bad' datasets with the previously removed EHR context data, which resulted in 113,659 `good' and 194,580 `bad' data points with (EHR context, medical jargon terms, lay definition, general definition) format. 
After eliminating duplicates, the final count for bad data points in R-exp\_bad was 177,140.

\section{General Definition Retrieval and Preprocessing}
\label{apx:umls_def}

UMLS~\cite{lindberg1993unified} is a set of files and software that combines many health and biomedical vocabularies and standards to enable interoperability between computer systems. 
Given the complexity of medical terminology, it's noteworthy that some terms in UMLS are associated with multiple definitions.
This reflects the reality that a medical term's meaning can vary depending on its contextual use.
To accurately select the most appropriate general definition for each context, we utilized the SentenceBERT~\cite{reimers2019sentence} similarity score between lay definitions and all possible UMLS definitions available for a jargon term to identify the most fitting definition.

Jargon terms in our dataset varied in length and composition, with not all words necessarily having corresponding UMLS definitions. This variability necessitated the development of distinct approaches for different scenarios:

\begin{enumerate}
[topsep=0.5pt,itemsep=0.2ex,partopsep=0.2ex,parsep=.20ex]
    \item \textbf{Single-word Terms:} In cases where the jargon term comprised a single word, we either found a UMLS definition or none. Data points lacking a UMLS definition in this category were excluded.
    \item \textbf{Two-word Terms:} For jargon terms composed of two words (word1 and word2), we considered several subcases:
    \begin{enumerate}
        \item Only word1 has a UMLS definition.
        \item Only word2 has a UMLS definition.
        \item Both word1 and word2 have UMLS definitions.
        \item The phrase ``word1 word2'' has a collective UMLS definition.
    \end{enumerate}
    \item \textbf{Terms with More than Two Words:} Similar to the case with two words, we can have cases where few words have UMLS definitions in the jargon term and few do not. 
\end{enumerate}

\begin{figure}
    \centering
    \includegraphics[width=0.75\linewidth]{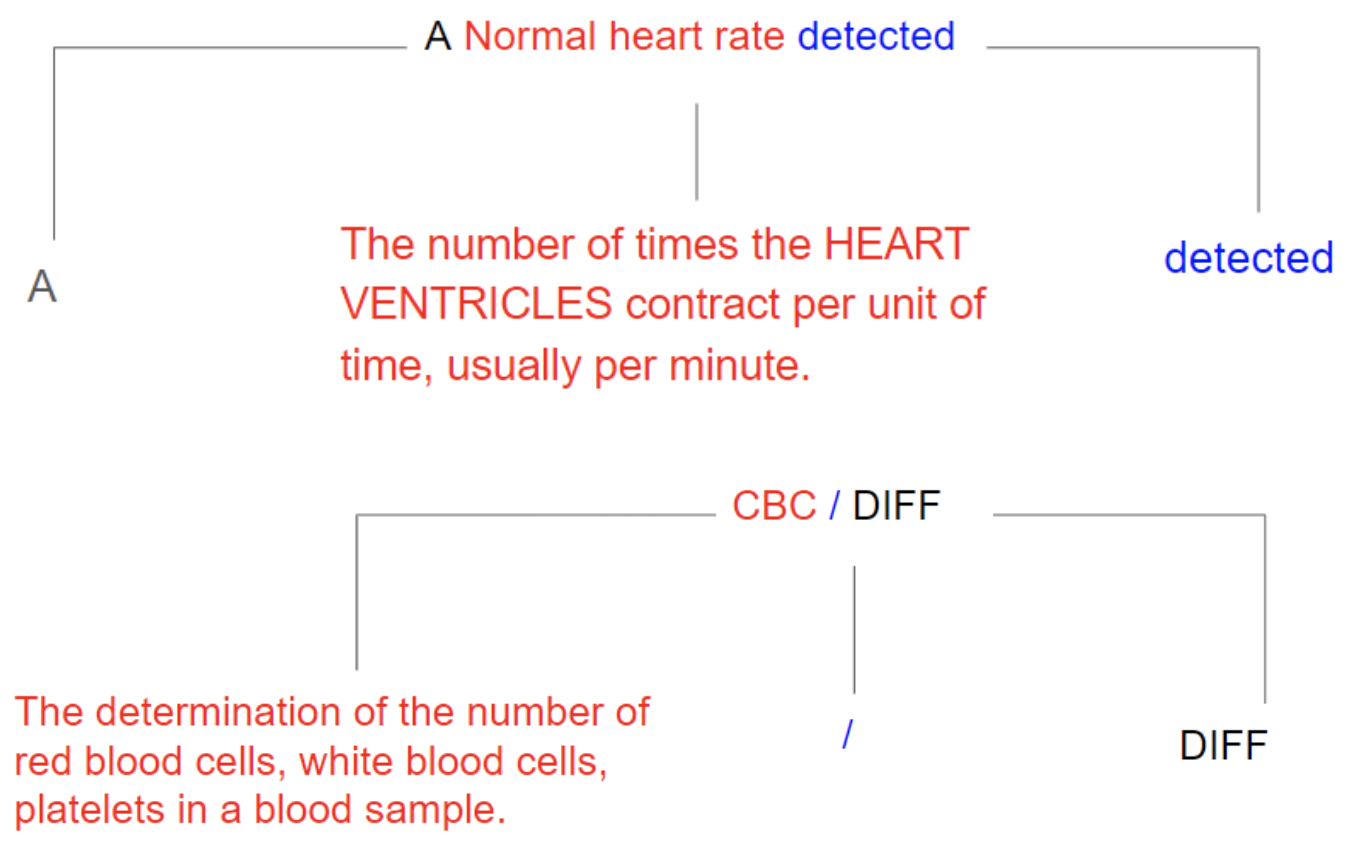}
    \caption{General definitions of two jargon examples.}
    \label{fig:general_definition}
    \vspace*{-5mm}
\end{figure}

Our solution to address these scenarios involved a unified approach, given the impracticality of tackling each case individually. As illustrated in Figure~\ref{fig:general_definition}, we extracted the UMLS definitions for all phrases within a jargon term. This process yielded a list of strings, providing a comprehensive general definition for each term. We then concatenate these strings in a comma-separated manner to get one string used as a general definition. There might not be a UMLS definition for all the words in the jargon term. In that case, we are concatenating the word directly. 

The initial README dataset has a lot of unusable data. There are many reasons why a data point might not be useful. The main reasons are (i) the lay definition provided by the annotator is not correct or missing. (ii) The jargon term is missing. (iii)The general definition obtained from UMLS is too scientific or not relevant to the jargon term. This was done by making sure there were no empty column values and removing parsing issues of comma-separated files.
Using this data for training would be bad for the training as these data points will make the model worse. 
So these data points are not considered. This preliminary cleaning brings out data from 350K to 308K.

\section{Discussion on Concept Ambiguity}
\label{apx:concept_ambiguity}

An important aspect that merits discussion is the potential ambiguity of medical concepts and its impact on generating lay user explanations. 
The inherent complexity and context-dependency of medical terms can create challenges in crafting universally understandable definitions, potentially leading to patient misinterpretations.
Our approach to addressing the ambiguity of medical concepts is rooted in the Word-Sense Disambiguation (WSD) phase, as elaborated in related works such as~\cite{kwon2022medjex}.
The WSD phase links ambiguous medical terms to accurate, disambiguated definitions from medical concept dictionaries.
These definitions serve as the foundation for generating lay definitions suitable for patient understanding, especially when dictionary definitions lack readability. In our study, the WSD phase is implicitly managed during the general definition retrieval stage using Scispacy and UMLS tools.

\section{Discussion on EAE Pipeline Efficacy}
\label{apx:eae_efficacy}

The primary objective of the EAE pipeline is to validate the effectiveness of Scispacy + UMLS for Word Sense Disambiguation (WSD), rather than addressing quality issues in the expert-annotated lay definitions. While quality checks and filtering for (jargon, context, lay definition) do not necessitate an LLM, they are essential for (jargon, context, lay definition, general definition) due to the noise introduced by Scispacy + UMLS tools. It is crucial to note that this noise is outside the scope of this paper.

An important aspect that merits discussion is the efficacy of the EAE pipeline design, particularly regarding the high number of data points categorized as R-exp\_bad or R-syn\_bad in Table~\ref{tab:dataset statistics}. 
Ideally, a more efficient pipeline would achieve a higher number of R-exp\_good data points, thereby reducing the need for additional verification rounds. 
However, our current settings prioritize reducing false positives, even if this results in an increased number of false negatives. 
Strict AI verification is essential to mitigate false positives, which could lead to patient misunderstandings if incorrect definitions are generated. 
Given the large size of the original README dataset, the harm of false negatives is more acceptable.

Therefore, we have intentionally set a stricter standard for AI during prompting (as shown in Table 9).
This approach may classify some good-quality data as bad, but it ensures that any data passing the Examiner stage is of high quality. 
This is corroborated by the high human agreement rates observed in Appendix~\ref{apx:soft_hard_correlation}.
Consequently, the final small human verification step is manageable without significantly increasing the workload.

The EAE pipeline and related prompts effectively detect and filter (jargon, context, lay definition, general definition) with minimal human effort, ensuring a sufficient quantity of valid data for subsequent training. 
In scenarios where the dataset is smaller or where more false positives are acceptable, the sensitivity of the AI examiner may need to be adjusted. 
Nevertheless, the three stages of the E→A→E pipeline are crucial for maintaining data quality in our task, as highlighted in Table~\ref{table:EAE_results}, and can be extended to other similar scenarios.

\section{Data Quality Checking and Train/Eval/Test Split after EAE Pipeline}
\label{apx:soft_hard_correlation}

After the end of the EAE, we use R-exp\_good and R-syn\_good as high-quality data for our system. 
The dataset was split into two categories: human examination data (which will also be used as final evaluation and test data since medical experts examine this split), and training data, where we ensure the medical jargon in the human examination will not appear in the training split. 
We sampled 500 medical jargon terms each from the R-exp\_good and R-syn\_good datasets.
Therefore the human examination split consisted of 1000 unique terms, each accompanied by general and lay definitions to be rated based on two criteria:

\begin{enumerate}
[topsep=0.5pt,itemsep=0.2ex,partopsep=0.2ex,parsep=.20ex]
    \item Hard Correlation: Marked `Yes' if the lay definition closely rephrases or shares significant wording with the general definition, implying comprehensibility without advanced medical knowledge.
    \item Soft Correlation: Marked `Yes' if the general definition accurately represents the term but is slightly contextually misaligned; marked `No' if the definition is incorrect or overly verbose, complicating the derivation of a lay definition.
\end{enumerate}

Here is one example for \textbf{Term} `von Willebrand disease':

\textbf{Expert Definition(lay definition)}: A bleeding disorder. It affects the blood's ability to clot

\textbf{General definition}: Hereditary or acquired coagulation disorder characterized by a qualitative or quantitative deficiency of the von Willebrand factor. The latter plays an important role in platelet adhesion. Signs and symptoms include bruises, nose bleeding, gum bleeding following a dental procedure, heavy menstrual bleeding, and gastrointestinal bleeding.

Although the lay definition is not incorrect, it is very wordy and complex and matches less to the lay definition. We consider this soft correlated and not hard correlated. A Hard Correlation automatically implies a Soft Correlation.

Two medical students~\footnote{Both have hospital internship experience} help us finish this human examination.
Our findings revealed that 88\% of R-exp\_good and 100\% of R-syn\_good met the Hard Correlation criteria. 
There is a 100\% soft Correlation criteria for both R-exp\_good and R-syn\_good. 
After correcting individual invalid data (e.g., those cases where Soft is not satisfied), we used this human examination dataset as evaluation and test data (in a 1:1 ratio).

\begin{table*}
\scalebox{0.95}{
\setlength{\tabcolsep}{3.9pt}
\renewcommand{\arraystretch}{1.05}
\vspace*{-3mm}
\centering
\begin{footnotesize}
\begin{tabularx}{1.0\textwidth}{X} 

\hline
In this task, we ask for your expertise in generating the corresponding lay definition from the medical jargon. Mainly, we provide the target medical jargon term. We need you to generate a lay definition for this jargon term.
\newline
\newline
Example:
\newline
jargon term: [TERM]
\newline
lay definition: [DEFINITION]
\newline
\newline
jargon term: [TERM]
\newline
lay definition:
\\
\hline
\end{tabularx}
\end{footnotesize}
}
\caption{One shot prompt for experiment set-1.}
\label{one_shot_prompt}
\end{table*}

\section{Factuality metrics: UMLS-F1}
\label{apx:umlsf1}

The assessment of factual accuracy in generated lay definition leverages the UMLS concept overlap metric. The Unified Medical Language System (UMLS), established by~\cite{bodenreider2004unified}, significantly contributes to the biomedical domain's interoperability. It achieves this by amalgamating and disseminating a comprehensive collection of biomedical terminologies, classification systems, and coding standards from many sources. By doing so, UMLS aids in reconciling semantic variances and representational disparities found across different biomedical concept repositories.

For the identification and alignment of medical named entities within texts to their corresponding biomedical concepts in UMLS, we employed the Scispacy library~\footnote{We used the Scispacy \textit{en\_core\_sci\_lg} model.}.
Scispacy excels in identifying and clarifying entities, thus facilitating the accurate association of named entities found in lay definitions with the relevant UMLS concepts. 
This capability is critical for evaluating the lay definitions' factual accuracy and is used by recent related work~\cite{adams2023meta}.

The analytical process for lay definitions utilizes metrics of precision and recall. Precision represents the ratio of concepts present in both the generated and reference lay definitions, serving as a measure of the generated lay definition's factual correctness. In contrast, recall evaluates how well the information in the generated lay definition matches the intended content, reflecting the relevance of the presented information.

To calculate these metrics, we consider the concept sets from both the reference lay definition ($C_{ref}$) and the generated lay definition ($C_{gen}$). The formulas for recall and precision are as follows:

$$
\text{Recall} = \frac{|C_{ref} \cap C_{gen}|}{|C_{ref}|}
$$

$$
\text{Precision} = \frac{|C_{ref} \cap C_{gen}|}{|C_{gen}|}.
$$

The F1 score, derived from the above precision and recall values, is reported to provide a balanced measure of the generated lay definition's accuracy and relevance.

\section{More Experimental Settings}
\label{apx:exp_setting}
We use the following base models: GPT-2 \footnote{\url{https://huggingface.co/gpt2}} \cite{radford2019language}, DistilGPT2 \footnote{\url{https://huggingface.co/distilgpt2}}, BioGPT \footnote{\url{https://huggingface.co/microsoft/biogpt}} \cite{10.1093/bib/bbac409}, and Llama2 \footnote{\url{https://huggingface.co/meta-llama/Llama-2-7b-chat-hf}} \cite{touvron2023llama} in our experiments.
We trained the base models on the different README dataset variants with Supervised Fine-tuning for 100000 steps (batch size 8) \footnote{We did all the experiments with 1 NVIDIA Tesla RTX 8000 GPU - 40 GB memory, with Adam optimizer -- betas=(0.9,0.999), epsilon=1e-08, learning rate=5e-04.}.
In all our evaluations, we used a beam size of 4, no-repeat-ngram-size=2, and minimum length and maximum length of sentences were set as (10, 100). We used five different random seeds to sample training data for all our experiments, and the scores reported in the tables are the average of these random seeds.

Here, we provide an example for jargon2lay(J2L), jargon+context2lay(J+C2L), jargon+gen2lay(J+G2L), and  jargon+context+gen2lay(J+C+G2L) for easier understanding. Let's assume we one data point in README dataset with (jargon, EHR context, lay def, general def) format:

\begin{itemize}
    \item \textbf{jargon:} EGD
    \item \textbf{lay def:} [esophagogastroduodenoscopy] A procedure that looks at the food pipe, stomach, and the first part of the small bowel.
    \item \textbf{EHR context:} [ * * 11 - 22 * * ] EGD Grade I varices - ablated [ * * 11 - 22 * * ] sigmoidoscopy friability , reythema , congest and abnormal vasularity in a small 5 mm area of distal rectum .
    \item \textbf{general def:} ['An endoscopic procedure that visualizes the upper part of the gastrointestinal tract up to the duodenum.']
\end{itemize}

The input of different settings J2L, J+C2L, J+G2L, J+C+G2L can be found on Table~\ref{different_settings_prompt}.

\section{More Experimental Designs and Results}
\label{apx:exp_design_result}
In Section~\ref{Sec:experiments}, our experimental design focuses on evaluating the quality of outputs generated by different systems. Here, ``quality'' is measured by two main criteria: the overall similarity between the system's output and the ground truth lay definition (e.g., ROUGE or METEOR), and the presence or absence of factual inaccuracies in the generated lay definition (UMLS-F1).
In this appendix section, we will focus more on readability.
Specifically, we follow the ReadCtrl~\cite{tran2024readctrl} to explore how the README dataset aids models in generating outputs with more controllable readability.
Therefore, we conducted instruction-following experiments with GPT-3.5 (few-shot), GPT-4 (few-shot), Claude3-opus (few-shot), Llama2-chat (few-shot), and Llama2-README-finetuning (few-shot). 
The prompt used was: 
\textit{``Given an input jargon term and general definition, please output a lay definition with a readability score around target readability [X].''}

\noindent where [X] was replaced by FKGL 1-12. A good instruction following should output lay definitions with readability scores similar to the target FKGL.

\begin{table}[h!]
    \centering
    \scalebox{0.8}{
    \begin{tabular}{|c|c|c|c|c|c|}
        \hline
        [X] & GPT-3.5 & GPT-4 & Claude3 & Llama2 & Ours \\
        \hline
        1  & 7.1410  & 6.0820 & 7.5364  & 10.4919 & 3.8800 \\
        2  & 6.7836  & 6.5907 & 7.2024  & 10.4365 & 4.5106 \\
        3  & 6.7412  & 7.4916 & 7.9721  & 10.4571 & 5.5185 \\
        4  & 7.5948  & 7.7300 & 8.4284  & 10.9103 & 6.1644 \\
        5  & 7.9722  & 8.1104 & 9.7814  & 10.6538 & 6.6462 \\
        6  & 8.7160  & 8.4608 & 10.9537 & 10.3240 & 6.9269 \\
        7  & 9.0761  & 8.9479 & 11.1111 & 10.3477 & 7.4499 \\
        8  & 10.0191 & 9.6390 & 13.3369 & 10.6044 & 8.2328 \\
        9  & 11.3319 & 11.0364 & 14.7280 & 10.5953 & 8.9487 \\
        10 & 12.4661 & 11.9267 & 16.5011 & 10.0969 & 9.5266 \\
        11 & 13.4467 & 12.4227 & 16.8663 & 10.6457 & 10.0348 \\
        12 & 13.2357 & 13.3720 & 17.2713 & 10.2263 & 10.5039 \\
        \hline
    \end{tabular}
    }
    \caption{Mean FKGL Scores for Each Model}
    \label{tab:fkgl-scores}
\end{table}

\begin{figure}
    \centering
    \vspace{-2mm}
    \includegraphics[width=1\linewidth]{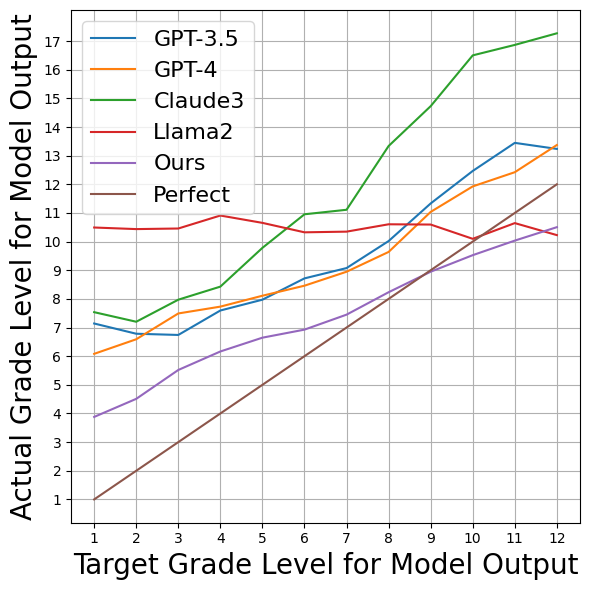}
    \caption{ReadCtrl~\cite{tran2024readctrl} instruction following ability using README dataset.}
    \label{fig:ReadCtrl}
\end{figure}

As illustrated in Table~\ref{tab:fkgl-scores} and Figure~\ref{fig:ReadCtrl}, our investigation across a range of
state-of-the-art LLMs shows varying degrees of
compliance with readability-controlled instructions. 
Mainstream models like GPT-3.5, GPT-4, and Claude3 are upward but far from the perfect curve, indicating they can follow instructions but not precisely. 
Llama2 does not show an upward trend, suggesting it cannot follow instructions. 
In contrast, Llama2-README closely follows the perfect curve, indicating precise instruction-following capability.

These results suggest that the README dataset contains sufficiently diverse readability information, making it highly useful for controllable text generation, particularly in readability control. This capability has significant potential for personalized patient education and represents a promising future research direction.

\begin{table*}
\scalebox{0.95}{
\setlength{\tabcolsep}{3.9pt}
\renewcommand{\arraystretch}{1.05}
\vspace*{-3mm}
\centering
\begin{footnotesize}
\begin{tabularx}{1.0\textwidth}{X} 

\hline
\textbf{jargon2lay(J2L):}
\newline
In this task, we ask for your expertise in generating the corresponding lay definition from the medical jargon. Mainly, we provide the target medical jargon term. We need you to generate a lay definition for this jargon term. 
\newline
jargon term: EGD 
\newline
lay definition:
\newline
\newline
\textbf{jargon+context2lay(J+C2L):}
\newline
In this task, we ask for your expertise in generating the corresponding lay definition from the medical jargon. Mainly, we provide the target medical jargon term along with the contextual snippets in which they appear in the text. We need you to generate a lay definition for this jargon term.
\newline
jargon term: EGD 
\newline
context: [ * * 11 - 22 * * ] EGD Grade I varices - ablated [ * * 11 - 22 * * ] sigmoidoscopy friability , reythema , congest and abnormal vasularity in a small 5 mm area of distal rectum .
\newline
lay definition:
\newline
\newline
\textbf{jargon+gen2lay(J+G2L):}
\newline
In this task, we ask for your expertise in generating the corresponding lay definition from the medical jargon. Mainly, we provide the target medical jargon term. In addition, we also provide a definition from the dictionary for reference. We need you to generate a lay definition for this jargon term.
\newline
jargon term: EGD 
\newline
dictionary definition: ['An endoscopic procedure that visualizes the upper part of the gastrointestinal tract up to the duodenum.']
\newline
lay definition:
\newline
\newline
\textbf{jargon+context+gen2lay(J+C+G2L):}
\newline
In this task, we ask for your expertise in generating the corresponding lay definition from the medical jargon. Mainly, we provide the target medical jargon term along with the contextual snippets in which they appear in the text. In addition, we also provide a definition from the dictionary for reference. We need you to generate a lay definition for this jargon term.
\newline
jargon term: EGD 
\newline
context: [ * * 11 - 22 * * ] EGD Grade I varices - ablated [ * * 11 - 22 * * ] sigmoidoscopy friability , reythema , congest and abnormal vasularity in a small 5 mm area of distal rectum .
\newline
dictionary definition: ['An endoscopic procedure that visualizes the upper part of the gastrointestinal tract up to the duodenum.']
\newline
lay definition:
\\
\hline
\end{tabularx}
\end{footnotesize}
}
\caption{The prompt of different settings J2L, J+C2L, J+G2L, J+C+G2L.}
\label{different_settings_prompt}
\end{table*}

\begin{algorithm}
  \caption{Algorithm for Data Cleaning}
  \label{algo:data_cleaning}
  \SetKwInOut{Input}{inputs}
  \SetKwInOut{Output}{output}
  \SetKwProg{READMEEAE}{README-EAE}{}{}

  \READMEEAE{}{
    \Input{README-exp}
    \Output{Final README dataset with best general definitions and lay definitions}
    $Initialize:$
    $exam_{p} = examiner prompt for gpt3$\;
    $aug_{p} = augmenter prompts for ChatGPT$\;
    \ForEach{$datapoint : README - exp$}{%
     \If{$exam_{p}(datapoint) == yes$}{
     R-exp\_good.add(datapoint)\;
     }
     \Else{
     R-exp\_bad.add(datapoint);
     }
    }
    \ForEach{$datapoint : R-exp\_bad$}{%
     $temp = aug_{p}(datapoint)$\;
     R-syn.add(temp)\;
     \If{$exam_{p}(temp) == yes$}{
     R-syn\_good.add(temp)\;
     }
     \Else{
     R-syn\_bad.add(temp)\;
     }
    }
    \KwRet{R-syn\_bad}\;
  }
\end{algorithm}

\onecolumn
\lstset{
    basicstyle=\ttfamily,
    columns=fullflexible,
    breaklines=true,
    postbreak=\mbox{\textcolor{red}{$\hookrightarrow$}\space}
}
{\tiny
\centering
\begin{tabular}{|p{15.6cm}|}
\hline
\textbf{Prompt} \\ \hline
\begin{lstlisting}
Examiner - 1

model = "gpt-3.5-turbo(ChatGPT API)"
[This examiner prompt does not use context as discussed in section 3.2.1]

Decide whether the general definition is correct.

If we can generate the lay definition from the general definition then answer is yes.

term : mg
general definition : this is short for milligram which is 1/1000 of a gram usually considered a small amount.
lay definition : A tiny amount of something, usually a drug.
answer : yes

term : vitamin c
general definition : [`A nutrient that the body needs in small amounts to function and stay healthy. Vitamin C helps fight infections, heal wounds, and keep tissues healthy. It is an antioxidant that helps prevent cell damage caused by free radicals (highly reactive chemicals). Vitamin C is found in all fruits and vegetables, especially citrus fruits, strawberries, cantaloupe, green peppers, tomatoes, broccoli, leafy greens, and potatoes. It is water-soluble (can dissolve in water) and must be taken in every day. Vitamin C is being studied in the prevention and treatment of some types of cancer.']
lay definition : A nutrient needed by the body to form and maintain bones, blood vessels, and skin.
answer : yes

term : nodule
general definition : [`A small lump, swelling or collection of tissue.']
lay definition : A growth or lump that may be cancerous or not.
answer : yes

term : qd
general definition : [`Occurring or done each day.']
lay definition : Every day.
answer : yes

If the general definition contains many words from the term then answer is no.

term : prochlorperzine 
general definition : [`prochlorperzine', ` ']
lay definition : A drug used to prevent or reduce nausea and vomiting.
answer : no

term : mg
general definition : [`mg']
lay definition : A tiny amount of something, usually a drug.
answer : no

If the lay definition can not be generated by the general definition then answer is no.

term : Virt - Vite
general definition : [`Virt', ` - ', `The determination of the amount of Vitamin E present in a sample.']
lay definition : A mix of vitamins. It provides vitamin B-6, vitamin B-12 and folic acid to people who do not have enough of these for good health.
answer : no
\end{lstlisting}
\\ \hline
 
\begin{lstlisting}
Augmenter

system_prompt = "your job is to generate a general definition of the term."
model="gpt-3.5-turbo(ChatGPT API)",
messages=[
        {"role": "system", "content": system_prompt},
        {"role": "user", "content": ""},
        {"role": "user", "content": "term : incisional."},
        {"role": "assistant", "content": "general definition : An intentional cut made to an individual's body with the intent of performing a diagnostic or therapeutic intervention."},
        {"role": "user", "content": "term : PO"},
        {"role": "assistant", "content": "general definition : Of, or relating to, or affecting, or for use in the mouth.."},
        {"role": "user", "content": prompt_t}
    ]
\end{lstlisting}
\\ \hline

\begin{lstlisting}
Examiner - 2

model = "gpt-3.5-turbo(ChatGPT API)"
exactly same as that of Examiner - 1
\end{lstlisting}
\\ \hline

\end{tabular}
}
\small
Table 9: All Examiner-Augmenter-Examine prompts.


\end{document}